\documentclass[runningheads]{llncs}
\usepackage{graphicx}
\usepackage{amsmath}
\usepackage{amssymb}
\usepackage{float}
\usepackage[utf8]{inputenc} 

\begin{document}
\title{Deep Eyes: Binocular Depth-from-Focus on Focal Stack Pairs}
%
%
\author{Xinqing Guo{*}\inst{2} \and
Zhang Chen{*}\inst{1} \and
Siyuan Li\inst{3} \and
Yang Yang\inst{2} \and
Jingyi Yu\inst{1}}
\authorrunning{X. Guo et al.}
%
\institute{
ShanghaiTech University, Shanghai, China\\
\email{\{chenzhang,yujingyi\}@shanghaitech.edu.cn}\and
DGene Inc., Santa Clara, USA\\
\email{\{xinqing,yyangwin\}@udel.edu}\and
École Polytechnique Fédérale de Lausanne, Lausanne, Switzerland\\
\email{siyuan.li@epfl.ch}
}

%
%
%
\maketitle              
\begin{abstract}
Human visual system relies on both binocular stereo cues and monocular
focusness cues to gain effective 3D perception. In computer vision,
the two problems are traditionally solved in separate tracks. In this
paper, we present a unified learning-based technique that simultaneously uses
both types of cues for depth inference. Specifically, we use a pair of
focal stacks as input to emulate human perception. We first construct
a comprehensive focal stack training dataset synthesized by
depth-guided light field rendering. We then construct three individual
networks: a \emph{Focus-Net} to extract depth from a single focal
stack, a \emph{EDoF-Net} to obtain the extended depth of field (EDoF)
image from the focal stack, and a \emph{Stereo-Net} to conduct stereo
matching. We show how to integrate them into a unified \emph{BDfF-Net} to
obtain high-quality depth maps. Comprehensive experiments show that
our approach outperforms the state-of-the-art in both accuracy and
speed and effectively emulates human vision systems. 

{\let\thefootnote\relax\footnote{{{*} These authors contribute to the work equally.}}}
\end{abstract}
\section{Introduction}
Human visual system relies on a variety of depth cues to gain 3D perception. The most important ones are binocular, defocus, and motion cues. Binocular cues such as stereopsis, eye convergence, and disparity yield depth from binocular vision through exploitation of parallax. Defocus cue allows depth perception even with a single eye by correlating variation of defocus blurs with the motion of the ciliary muscles surrounding the lens. Motion parallax also provides useful input to assess depth, but arrives over time and depends on texture gradients.

Computer vision algorithms such as stereo matching \cite{scharstein02} and depth-from-focus \cite{nayar92,malik07} seek to employ binocular and defocus cues which are available instantaneously without scene statistics. Recent studies have shown that the two types of cues complement each other to provide 3D perception \cite{held12}. In this paper, we seek to develop learning-based approaches to emulate this process.

To exploit binocular cues, traditional stereo matching algorithms rely on feature matching and optimization to maintain the Markov Random Field property. In contrast, depth-from-focus (DfF) exploits differentiations of sharpness at each pixel across a focal stack and assigns the layer with the highest sharpness as its depth. Compared with stereo, DfF generally presents a low fidelity estimation due to limited aperture size. Earlier DfF techniques use a focal sweep camera to produce a coarse focal stack due to mechanical limitations whereas more recent ones attempt to use a light field to synthetically produce a denser focal stack.

Our solution benefits from recent advances on computational photography and we present an efficient and reliable learning-based technique to conduct depth inference from a focal stack pair, emulating the process of how human eyes work. We call our technique binocular DfF or B-DfF. Our approach leverages deep learning techniques that can effectively extract features learned from large amount of imagery data. Such a deep representation has shown great promise in stereo matching \cite{zbontar15,luo16}. Little work, however, has been proposed on using deep learning for DfF or more importantly, integrating stereo and DfF. This is mainly due to the lack of fully annotated DfF datasets.

\begin{figure}[t]
\begin{center}
   \includegraphics[width=0.8\linewidth]{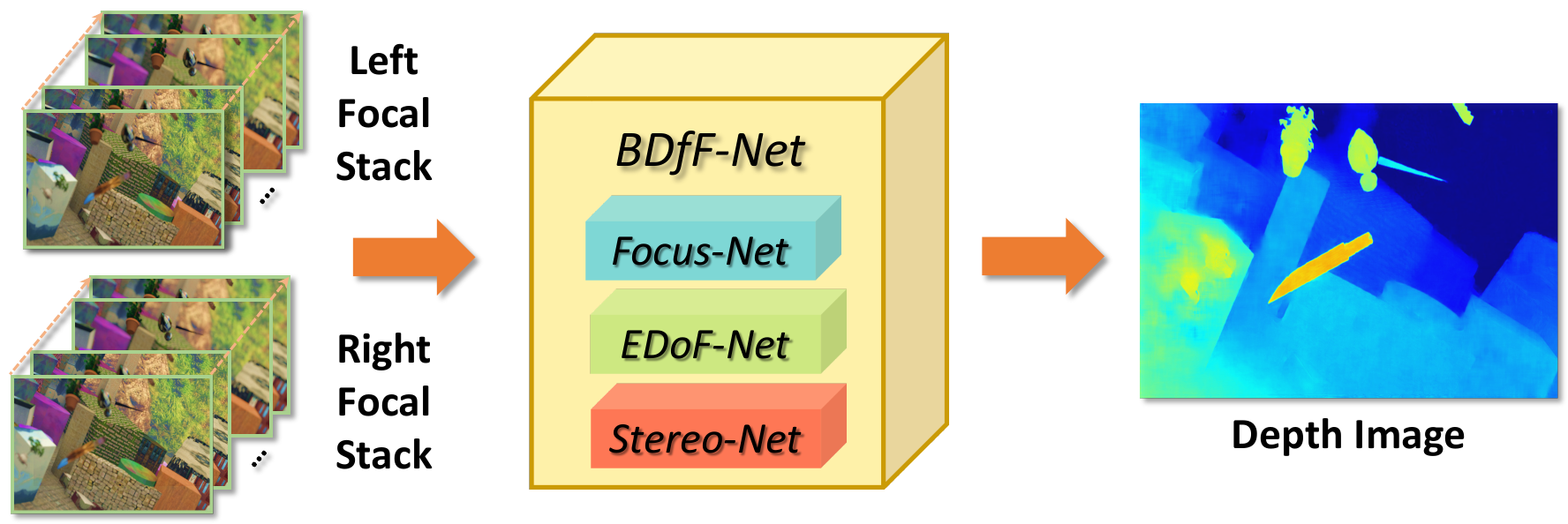}
\end{center}
\vspace{-8pt}
\setlength{\belowcaptionskip}{-10pt}
   \caption{\emph{BDfF-Net} integrates \emph{Focus-Net}, \emph{EDoF-Net} and \emph{Stereo-Net} to predict high quality depth map from binocular focal stacks.}
\label{fig:teaser}
\end{figure}

We first construct a comprehensive focal stack dataset. Our dataset is based on the highly diversified dataset from \cite{mayer16}, which contains both stereo color images and ground truth disparity maps. Then we adopt the algorithm from \emph{Virtual DSLR} \cite{yang16} to generate the refocused images. \cite{yang16} uses color and depth image pair as input for light field synthesis and rendering, but without the need to actually create the light field. The quality of the rendered focal stacks is comparable to those captured by expensive DSLR camera. Next, we propose three individual networks: (1) \emph{Focus-Net}, a multi-scale network to extract depth from a single focal stack (2) \emph{EDoF-Net}, a deep network consisting of small convolution kernels to obtain the extended depth of field (EDoF) image from the focal stack and (3) \emph{Stereo-Net} to obtain depth directly from a stereo pair. The EDoF image from \emph{EDoF-Net} serves to both guide the refinement of the depth from \emph{Focus-Net} and provide inputs for \emph{Stereo-Net}. We also show how to integrate them into a unified \emph{BDfF-Net} to obtain high-quality depth maps. Fig. \ref{fig:teaser} illustrates the pipeline.


We evaluate our approach on both synthetic and real data. To physically implement B-DfF, we construct a light field stereo pair by using two Lytro Illum cameras. Light field rendering is then applied to produce the two focal stacks as input to our framework. Comprehensive experiments show that our technique outperforms the state-of-the-art techniques in both accuracy and speed. 


\section{Related Work}
Our work is closely related to depth from focus/defocus and stereo. The strength and weakness of the two approaches have been extensively discussed in \cite{schechner00,vaish06}.

\noindent\textbf{Depth from Focus}
Blur carries information about the object's distance. Depth from Focus (DfF) recovers scene depth from a collection of images captured under varying focus settings. In general, DfF \cite{nayar92,malik07} determines the depth by analyzing the most in-focus slice in the focal stack. \cite{hasinoff09} combined focal stack with varying aperture to recover scene geometry. Suwajanakorn \emph{et al.} \cite{suwajanakorn15} proposed the DfF with mobile phone under uncalibrated setting. They first aligned the focal stack, then jointly optimized the camera parameters and depth map, and further refined the depth map using anisotropic regularization.

A drastic difference of these methods to our approach is that they rely on hand-crafted features to estimate the focusness, whereas in this paper we leverage the neural network to learn more discriminative features from the focal stack and directly predict the depth at a fraction of the computational cost.

\noindent \textbf{Learning based Stereo}
Depth from stereo has been studied extensively by the computer vision community for decades \cite{scharstein02}. Here we only discuss recent methods based on Convolutional Neural Network (CNN).

Deep learning benefits stereo matching at various stages. A number of approaches exploit CNN to improve the matching cost. The seminal work by {\v Z}bontar and LeCun~\cite{zbontar15} computed a similarity score from patches using CNN, then applied the traditional cost aggregation and optimization to solve the energy function. Luo \emph{et al.} \cite{luo16} speeded up the matching process by using a product layer, and treated the disparity estimation as a multi-class classification problem. ~\cite{chen2015deep,zagoruyko15,park2016look} conducted similar work but with different network architecture.

End-to-end network architectures have also been explored. Mayer \emph{et al.}~\cite{mayer16} adopted and extended the architecture of the FlowNet, which consists of a contractive part and an expanding part to learn depth at multiple scales. They also created three synthetic datasets to facilitate the training process. Kn{\"o}belreiter \emph{et al.}~\cite{knobelreiter2017end} learned unary and pairwise cost of stereo using CNNs, then posed the optimization as a conditional random field (CRF) problem. The hybrid CNN-CRF model was trained in image's full resolution in an end-to-end fashion.

Combining DfF and stereo matching has also been studied, although not within the learning framework. Early work~\cite{klarquist95} attempted to utilize the depth map from the focus to reduce the search space for stereo and solve the correspondence problem more efficiently. \cite{Rajagopalan04} simultaneously recovered depth and restored the original focused image from a defocused stereo pair.

Aforementioned approaches leave the combination and optimization of focus and disparity cue to post-processing. In contrast, we resort to extra layers of network to infer the optimized depth with low computational cost and efficiency.


\begin{figure}[t]
\begin{center}
   \includegraphics[width=0.81\linewidth]{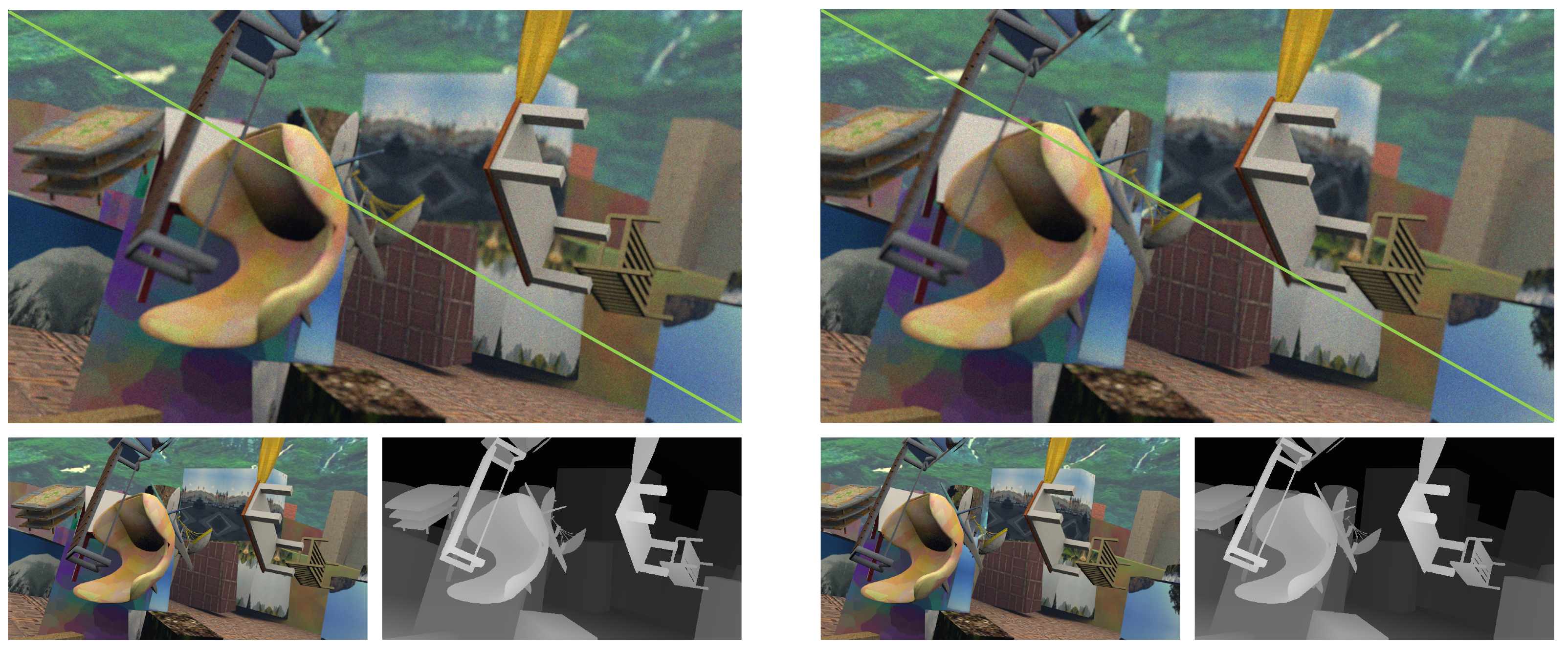}
\end{center}
\vspace{-8pt}
\setlength{\belowcaptionskip}{-10pt}
   \caption{A binocular focal stack pair consists of two horizontally rectified focal stacks. The upper and lower triangles show corresponding slices focusing at respective depths. Bottom shows the ground truth color and depth images. We add Poisson noise to training data, a critical step for handling real scenes.}
\label{fig:dataset_sample}
\end{figure}

\section{Dual Focal Stack Dataset}
With fast advances of the data-driven methods, numerous datasets have been created for various applications. However, by far, there are limited resources on focal stacks. To this end, we generate our dual focal stack dataset based on FlyingThings3D from \cite{mayer16}. Their 3D models and textures are separated into disjointed training and testing parts. In total, the dataset contains about 25,000 stereo images with ground truth disparity. To make the data tractable, we select stereo frames whose largest disparity is less than 100 pixels to avoid objects appearing in one image but not in the other.

Takeda \emph{et al.} \cite{takeda2013fusing} demonstrate that in a stereo setup, the disparity and the diameter of the circle of confusion have a linear relationship. Based on above observation, we adopt the \emph{Virtual DSLR} approach from \cite{yang16} to generate synthetic focal stacks. \emph{Virtual DSLR} requires color and disparity image pair as inputs, and outputs refocused images with quality comparable to those captured from regular, expensive DSLR. The advantage of their algorithm is that it resembles light field synthesis and refocusing but does not require actual creation of the light field, hence reducing both memory and computational load. In addition, their method takes special care of occlusion boundaries to avoid color bleeding and discontinuity commonly observed in brute-force blur-based defocus synthesis.

For binocular focal stack dataset, we evenly separate the scene into 16 depth layers and render a refocused image for each layer. Figure \ref{fig:dataset_sample} shows two slices from the dual focal stack and their corresponding color and depth images. We further add Poisson noise to both datasets to simulate real images captured by a camera. This turns out to be critical in real scene experiments, as described in section \ref{section:experiments}. Our final datasets each contain 750 training samples and 160 testing samples, with each sample consisting of 16 differently focused stereo image pair. The resolution of the generated images is $960\times540$, the same as the ones in FlyingThings3D.

\section{B-DfF Network Architecture}
\label{section:NetworkArchitectures}


When designing our network, one general principle is to use deep architecture with small kernels. \cite{Simonyan2015VeryDC} shows that such a structure is very effective in image recognition tasks. Further, we aim to take an end-to-end approach to predict a depth map.

As already mentioned, the input to the neural network is two rectified focal stacks. To extract depth from defocus and disparity, our solution is composed of three individual networks. We start in section \ref{section:multiscale} by describing the \emph{Focus-Net-Guided}, a multi-scale network that estimates depth from a single focal stack and further enhanced by the extended depth of field images from \emph{EDoF-Net}. Then we combine \emph{Stereo-Net} and \emph{Focus-Net-Guided} in \ref{section:stereo} to construct \emph{BDfF-Net} for high quality depth from binocular focal stacks.

\subsection{Focus-Net and Focus-Net-Guided for DfF}
\label{section:multiscale}

\begin{figure}[t]
\begin{center}
   \includegraphics[width=0.95\linewidth]{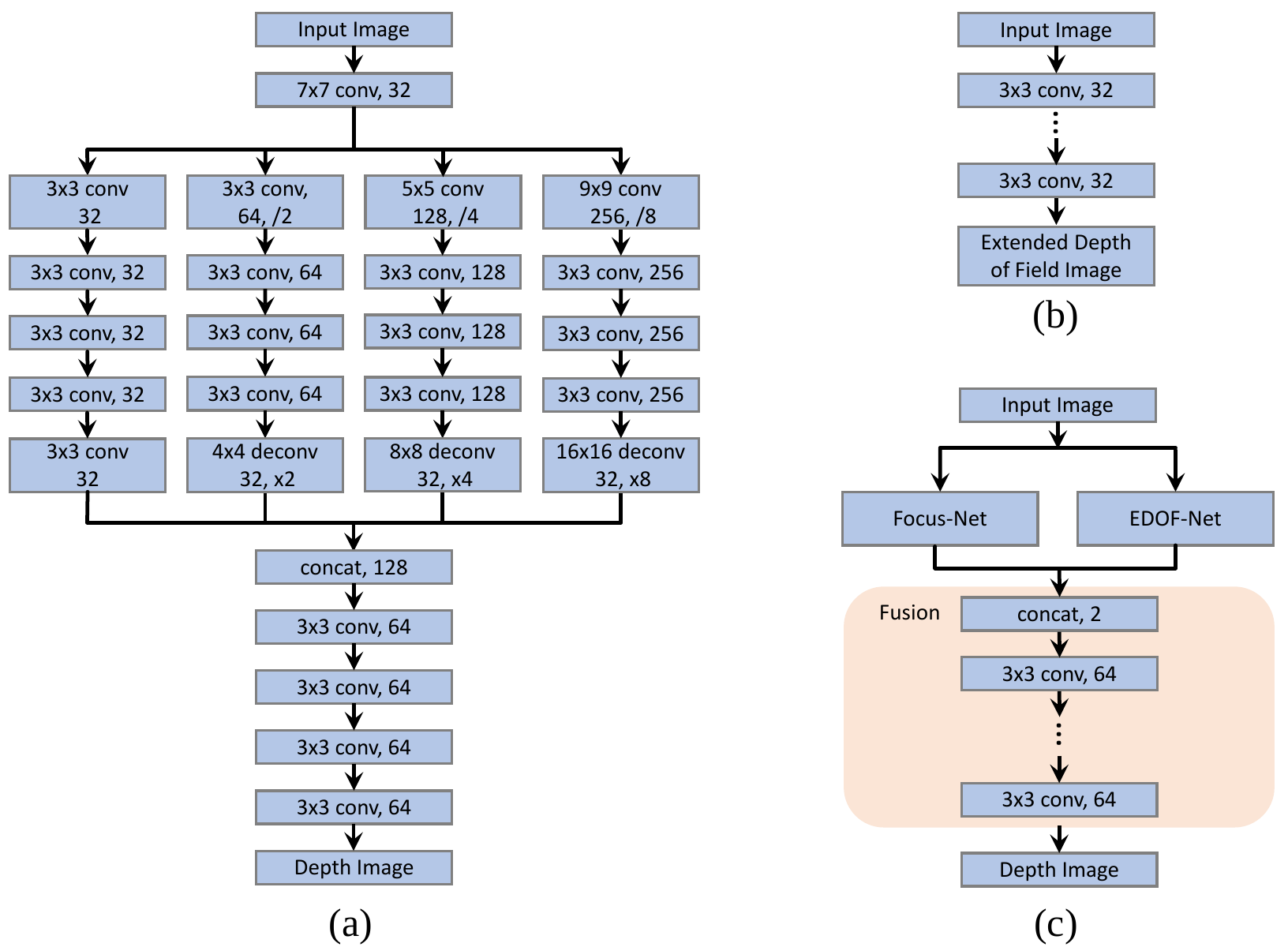}
\end{center}
\vspace{-8pt}
\setlength{\belowcaptionskip}{-10pt}
   \caption{(a) \emph{Focus-Net} is a multi-scale network for conducting depth-from-focus. (b) \emph{EDoF-Net} consists of 20 layers of convolutional layers to form an extended depth-of-field (EDoF) image from focal stack. (c) Our \emph{Focus-Net-Guided} combines \emph{Focus-Net} and \emph{EDoF-Net} by using the EDoF image to refine the depth estimation.}
\label{fig:network_FocusNet-v2}
\end{figure}

Motivated by successes from multi-scale networks, we propose \emph{Focus-Net}, a multiscale network to extract depth from a single focal stack. Specifically, \emph{Focus-Net} consists of four branches of various scales. Except for the first branch, other branches subsample the image by using different strides in the convolutional layer, enabling aggregation of information over large areas. Therefore, both the high-level information from the coarse feature maps and the fine details could be preserved. At the end of the branch, a deconvolutional layer is introduced to upsample the image to its original resolution. Finally, we stack the multi-scale features maps together, resulting in a concatenated per-pixel feature vector. The feature vectors are further fused by layers of convolutional networks to predict the final depth value.

An illustration of the network architecture is shown in Fig. \ref{fig:network_FocusNet-v2}(a). We use $3\times3$ kernels for most layers except those convolutional layers used for downsampling and upsampling, where a larger kernel is used to cover more pixels. Following \cite{Simonyan2015VeryDC}, the number of feature maps increases as the image resolution decreases. Between the convolutional layers we insert PReLU layer \cite{he15} to increase the network's nonlinearity. For the input of the network, we simply stack the focal stack images together along the channel's dimension.

To further enhance the depth image quality, we set out to extract the EDoF image from the focal stack, and use it to guide the refinement of the depth image. Existing EDoF image extraction methods \cite{kuthirummal11,suwajanakorn15} are suboptimal in terms of computational efficiency. Thus, we seek to directly output an EDoF image from a separate network, which we termed \emph{EDoF-Net}. \emph{EDoF-Net} is composed of 20 convolutional layers, with PRelu as its activation function. The input of the \emph{EDoF-Net} is the focal stack, the same as the input of \emph{Focus-Net}, and the output is the EDoF image. With the kernel size of $3\times3$, a 20 layer convolutional network will produce a receptive field of $41\times41$, which is larger than the size of the largest blur kernel. Fig. \ref{fig:network_FocusNet-v2}(b) shows the architecture of \emph{EDoF-Net}.

Finally, we concatenate the depth image from \emph{Focus-Net} and the EDoF image from the \emph{EDoF-Net}, and fuse them by using another 10 convolutional layers. We call the new network \emph{Focus-Net-Guided}, as illustrated in Fig. \ref{fig:network_FocusNet-v2}(c).


\subsection{Stereo-Net and BDfF-Net for Depth from Binocular Focal Stack}
\label{section:stereo}
Given the EDoF stereo pair from the \emph{EDoF-Net}, we set out to estimate depth from stereo using another network, termed \emph{Stereo-Net}. For stereo matching, it is critical to consolidate both local and global cues to generate precise pixel-wise disparity.

\begin{figure}[t]
\begin{center}
   \includegraphics[width=0.9\linewidth]{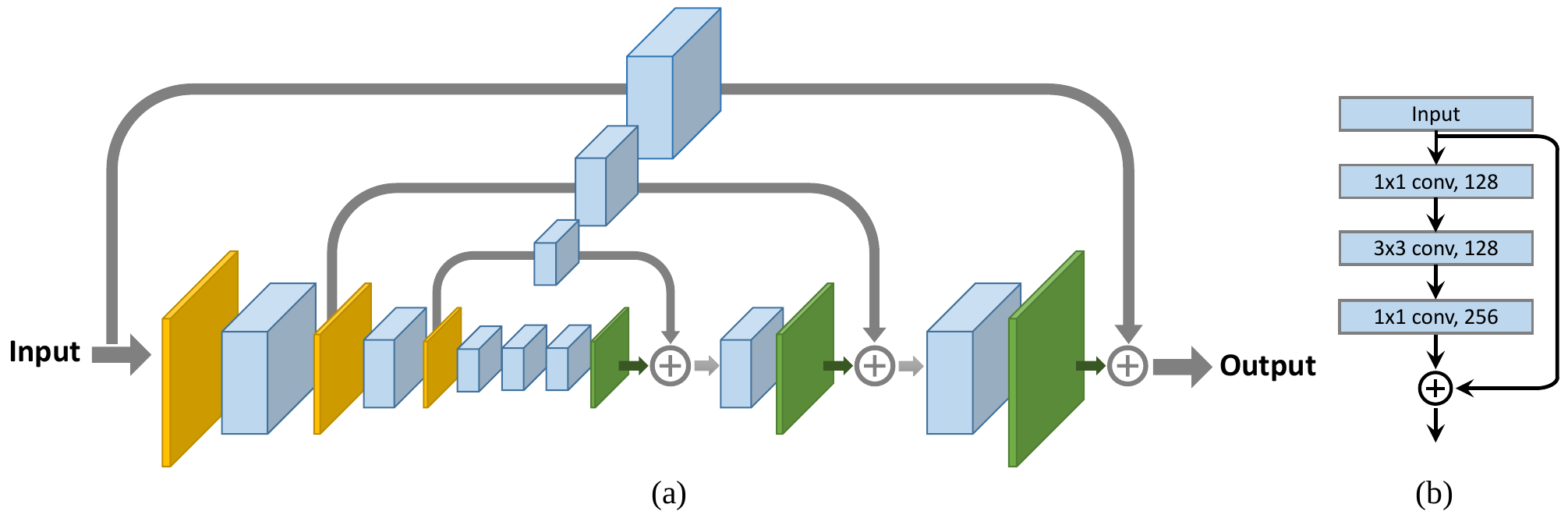}
\end{center}
\vspace{-8pt}
\setlength{\belowcaptionskip}{-10pt}
   \caption{(a) \emph{Stereo-Net} follows the Hourglass network architecture which consists of the max pooling layer (yellow), the deconvolution layer (green) and the residual module (blue). (b) shows the detailed residual module.}
\label{fig:network_hourglass}
\end{figure}

To this end, we propose \emph{Stereo-Net} by adopting the Hourglass network architecture~\cite{newell2016stacked}, as shown in Fig. \ref{fig:network_hourglass}. The advantage of this network is that it can attentively evaluate the coherence of features across scales by utilizing large amount of residual modules~\cite{he2016deep}. The network composes of a downsampling part, an upsampling part and connection layers comprising of residual modules. In this way, the network could both learn a holistic representation of input images and maintain fine structures. Prediction is generated at the end of the upsampling part. One round of downsampling and upsampling part can be viewed as one iteration of predicting, whereas additional rounds can be stacked to refine initial estimates. For \emph{Stereo-Net}, we use two rounds of downsampling and upsampling parts as they already give a good performance. After each pair of downsampling and upsampling parts, supervision is applied using the same ground truth disparity map. The final output is of the same resolution as the input images.

Finally, we construct \emph{BDfF-Net} by concatenating the results from \emph{Stereo-Net}, \emph{Focus-Net-Guided}, and adding 10 more convolutional layers. The convolutional layers serve to find the optimal combination from focus cue and disparity cue.

\section{Implementation}
\label{section:implementation}

Given the focal stack as input and ground truth color/depth image as label, we train all the networks end-to-end. In our implementation, we first train each network individually, then fine-tune the concatenated network with the pre-trained weights as initialization. Because \emph{Focus-Net} and \emph{Focus-Net-Guided} contains multiple convolutional layers for downsampling, the input image needs to be cropped to the nearest number that is multiple of 8 for both height and width. We use the mean absolute error with $l_2$-norm regularization as the loss for all models.

Following \cite{ioffe15}, we apply batch normalization after the convolution layer and before PRelu layer. We initialize the weights using the technique from \cite{he15}. We employ MXNET as the learning framework and train and test the networks on a NVIDIA K80 graphic card. We make use of the Adam optimizer and set the weight decay = 0.002, $\beta1$ = 0.9, $\beta2$ = 0.999. The initial learning rate is set to be 0.001. All the networks are trained for 80 epochs.


\section{Experiments}
\label{section:experiments}
\subsection{Extract the EDoF Image from Focal Stack}
\begin{figure}[t]
\begin{center}
   \includegraphics[width=1\linewidth]{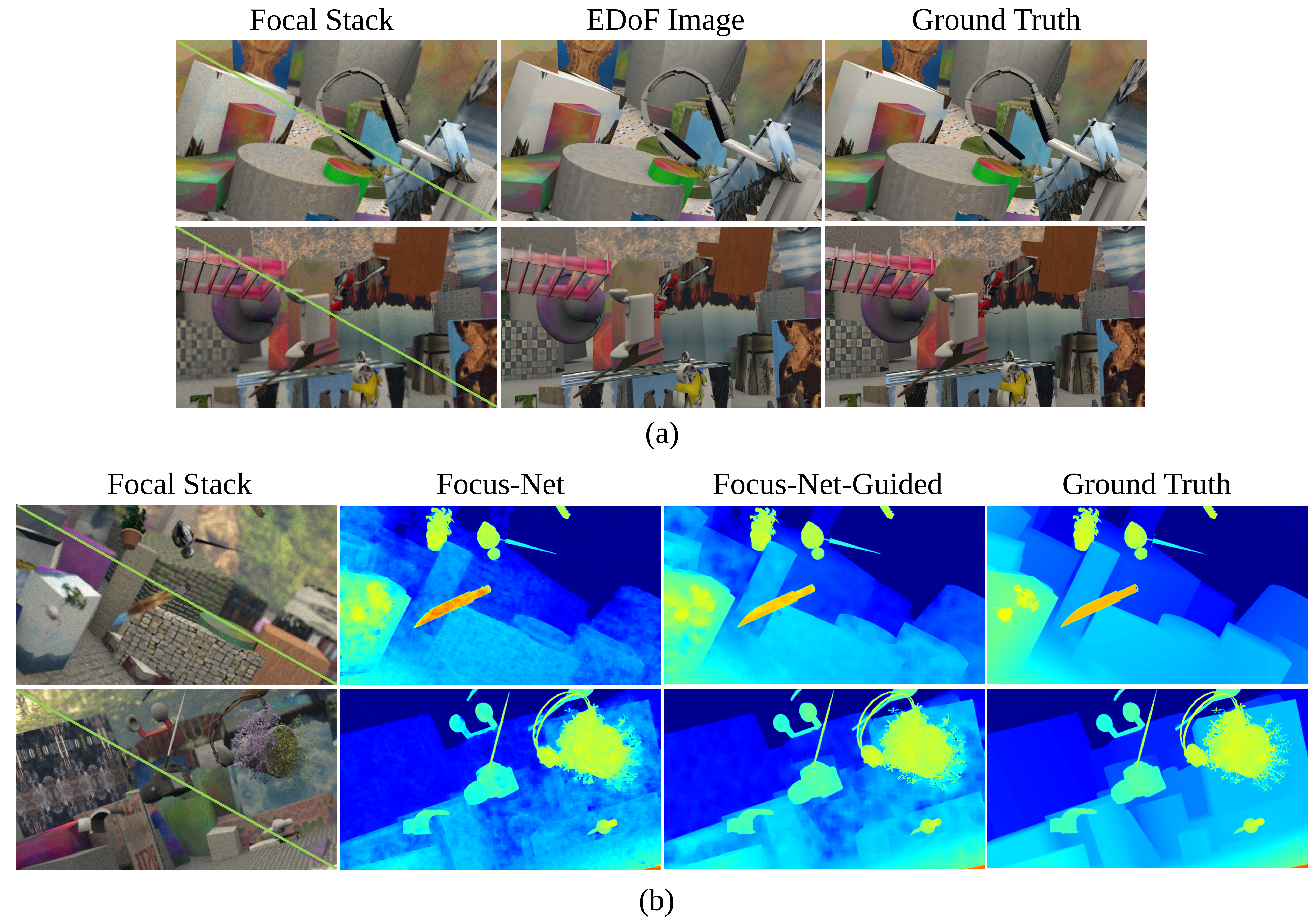}
\end{center}
\vspace{-8pt}
\setlength{\belowcaptionskip}{-10pt}
   \caption{(a) Results of our \emph{EDoF-Net}. First column shows two slices of the focal stack focusing at different depth. Second and third columns show the EDoF and ground truth image respectively. (b) Comparisons on \emph{Focus-Net} vs. \emph{Focus-Net-Guided}, i.e., without and with the guide of an all-focus image.}
\label{fig:result_DfF}
\end{figure}

We train \emph{EDoF-Net} on a single focal stack of 16 slices. Although the network has a simple structure, the output EDoF image features high image quality. Our network also runs much faster than conventional methods based on global optimization: for $960\times540$ images it runs at 4 frames per second. Fig. \ref{fig:result_DfF}(a) shows the result of \emph{EDoF-Net}. Our experiments also show that it suffices to guide the refinement of depth image and be used as the input of \emph{Stereo-Net}.


\subsection{Depth Estimation from Focal Stack}
\label{section:result_FocalStack}

As mentioned in \ref{section:multiscale}, to construct \emph{Focus-Net-Guided}, we first train \emph{Focus-Net} and \emph{EDoF-Net} respectively, then concatenate their output with more fusion layers and train the combination. Fig. \ref{fig:result_DfF} shows the result of both \emph{Focus-Net}
and \emph{Focus-Net-Guided}. We observe that \emph{Focus-Net} produces results with splotchy artifact, and depth bleeds across object's boundary. However,
\emph{Focus-Net-Guided} utilizes the EDoF color image to assist depth refinement, alleviating the artifacts and leading to clearer
depth boundary. It is worth noting that we also trained a network that
has identical structure to \emph{Focus-Net-Guided} from scratch, but the result is of inferior quality. We suspect this is due to the good initialization provided by the pre-trained model.

\begin{figure}[t]
\begin{center}
   \includegraphics[width=0.76\linewidth]{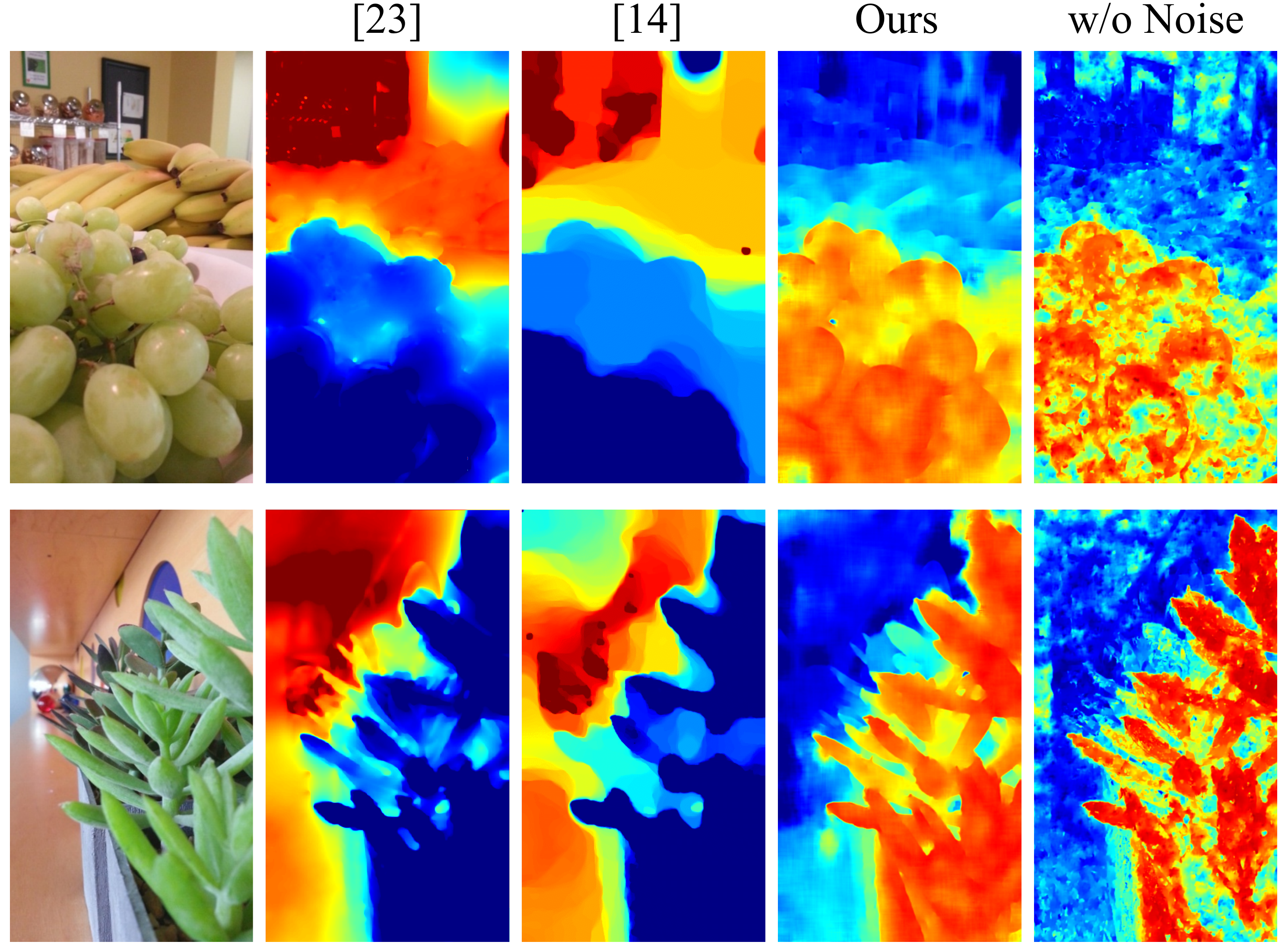}
\end{center}
\vspace{-8pt}
\setlength{\belowcaptionskip}{-10pt}
   \caption{Comparisons on depth estimation from a single focal stack using our \emph{Focus-Net-Guided} vs. \cite{suwajanakorn15} and \cite{moeller15}. \emph{Focus-Net-Guided} is able to maintain smoothness on flat regions while preserving sharp occlusion boundaries. The last column shows Results from \emph{Focus-Net-Guided} trained by the clean dataset without poisson noise.  \cite{suwajanakorn15} and \cite{moeller15} generate depth value while our \emph{Focus-Net-Guided} generates disparity value, so the colors of the images are inverted. }
\label{fig:result_compareVariation_v3}
\end{figure}

We compare our DfF results with \cite{suwajanakorn15} and \cite{moeller15} using the data provided by the authors of \cite{suwajanakorn15}. We select 16 images from their focal stack for DfF. Fig. \ref{fig:result_compareVariation_v3} illustrates the results. Our \emph{Focus-Net-Guided} is capable of predicting disparity value with higher quality, while using significantly less time (0.9 second) than \cite{suwajanakorn15} (10 minutes) and \cite{moeller15} (4 seconds).

We also train the \emph{Focus-Net-Guided} on a clean dataset without Poisson noise. It performs better on synthetic data, but exhibits severe noise pattern on real images, as shown in the last column of Fig. \ref{fig:result_compareVariation_v3}. The experiment confirms the necessity to add noise to the dataset for simulating real images.

\subsection{Depth Estimation from Stereo and Binocular Focal Stack}
\label{section:stereoAndBDfF}

Figure \ref{fig:result_BDfF} shows the results from \emph{Stereo-Net} and \emph{BDfF-Net}. Compared with \emph{Focus-Net-Guided}, \emph{Stereo-Net} gives better depth estimation. This is expected since \emph{Stereo-Net} requires binocular focal stacks as input, while \emph{Focus-Net-Guided} only use a single focal stack. However, \emph{Stereo-Net} exhibits blocky artifacts and overly smoothes boundary. In contrast, depth prediction from \emph{BDfF-Net} features sharper edges.
Table \ref{tab:result} describes the mean absolute error (MAE) and running time of all models.

\begin{table}[H]
\begin{center}
 \begin{tabular}{|c|c |c |c |c|}
    \hline
    \ & \emph{Focus-Net}   & \emph{Focus-Net-Guided} & \emph{Stereo-Net} & \emph{BDfF-Net} \\ [0.5ex]
    \hline
    MAE & 0.045  & 0.031 & 0.024 & 0.021 \\
    Time (s) & 0.6  & 0.9 & 2.8 & 9.7 \\
 \hline
\end{tabular}
\setlength{\belowcaptionskip}{-10pt}
\caption{MAE and running time of models.}
\label{tab:result}
\end{center}
\end{table}


\begin{figure}[t]
\begin{center}
   \includegraphics[width=1.0\linewidth]{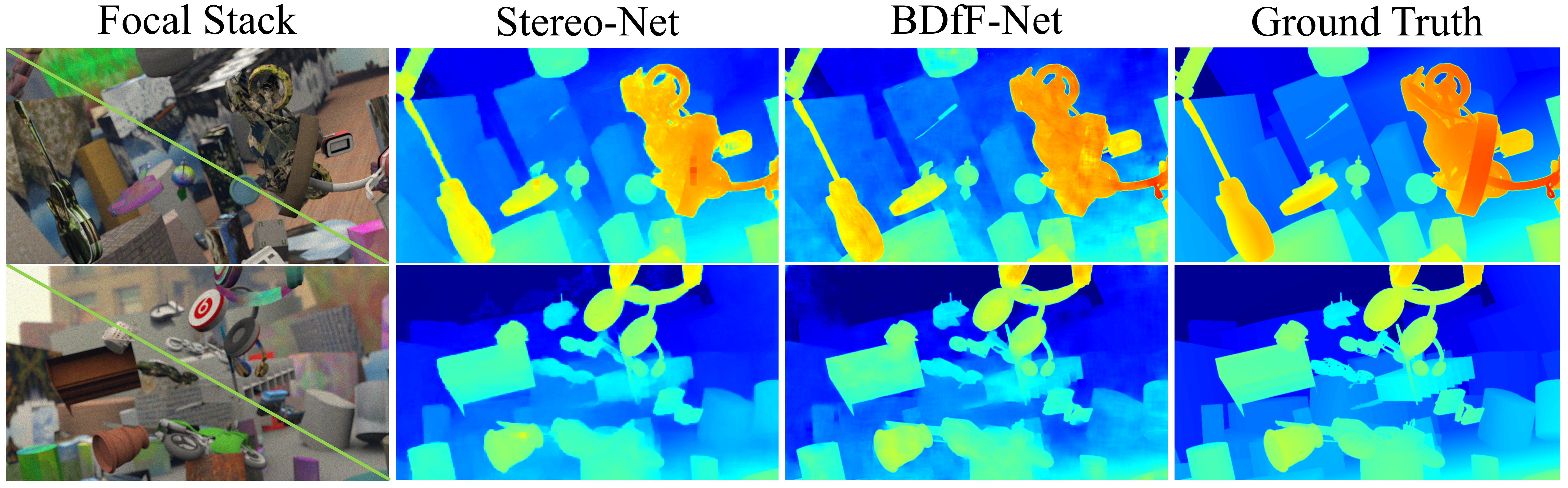}
\end{center}
\vspace{-8pt}
\setlength{\belowcaptionskip}{-5pt}
   \caption{Comparisons on results only using \emph{Stereo-Net} vs. the composed \emph{BDfF-Net}. \emph{BDfF-Net} produces much sharper boundaries while reducing blocky artifacts.}
\label{fig:result_BDfF}
\end{figure}

\begin{figure}[H]
\begin{center}
   \includegraphics[width=1.0\linewidth]{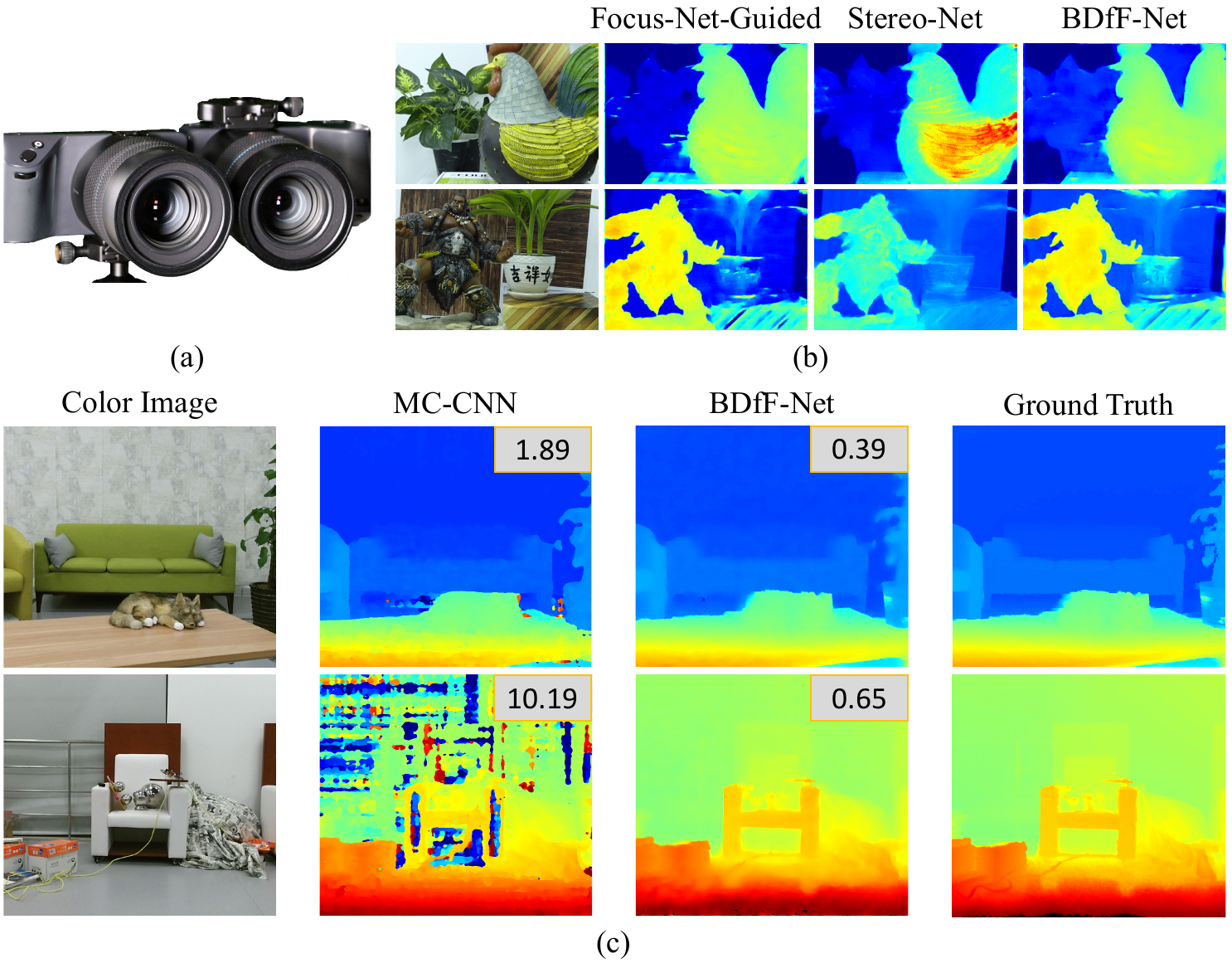}
\end{center}
\vspace{-8pt}
\setlength{\belowcaptionskip}{-10pt}
   \caption{(a) To emulate our B-DfF setup, we combine a pair of Lytro Illum cameras into a stereo setup. (b) Comparisons of \emph{Focus-Net-Guided}, \emph{Stereo-Net} and \emph{BDfF-Net} on data captured with (a). (c) Comparisons with \cite{zbontar2016stereo} on data captured with RGB-D camera (on top-right shows MAE of each predicted disparity map).}
\label{fig:result_realScene}
\end{figure}

\subsection{Real Scene Experiment}
\label{section:realScene}
We further conduct tests on real scenes. To physically implement B-DfF, we construct a light field stereo pair by using two Lytro Illum cameras, as illustrated in Fig. \ref{fig:result_realScene}(a). Light field camera contains a microlens array to capture multiple views of the scene, allowing users to perform post-capture refocusing. In our experiment, the two light field cameras share the same configuration including the zoom and focus settings. The raw images are pre-processed using Light Field Toolbox \cite{dansereau13}. Finally, we conduct refocusing using shift-and-add algorithm \cite{ng05} to synthesize the focal stack.
Figure \ref{fig:result_realScene}(b) shows the predicted depth from \emph{Focus-Net-Guided}, \emph{Stereo-Net} and \emph{BDfF-Net}. Results show that \emph{BDfF-Net} benefits from both \emph{Focus-Net-Guided} and \emph{Stereo-Net} to offer smoother depth with sharp edges. The experiments also demonstrate that models learned from our dataset could be transferred to predict real scene depth.

For quantitative analysis, we use a RGB-D camera (Kinect) to collect ground-truth depth. We mount the Kinect on a translation stage and move it horizontally to obtain a stereo pair of color images and disparity images, which we utilize to synthesize dual focal stacks using \emph{Virtual DSLR} \cite{yang16}.
Fig.\ref{fig:result_realScene}(c) compares our \emph{BDfF-Net} with the stereo matching method from \cite{zbontar2016stereo}. Note that our method produces accurate results in textureless regions while the results from \cite{zbontar2016stereo} contain large errors. 
This demonstrates the advantage of our approach, which effectively incorporates both focus and disparity cues in a multi-scale scheme.

\section{Discussions and Future Work}
\label{section:conclusion}

Our deepeye solution exploits efficient learning and computational light field imaging to infer depths from a focal stack pair. Our technique mimics human vision system that simultaneously employs binocular stereo matching and monocular depth-from-focus. Comprehensive experiments show that our technique is able to produce high-quality depth estimation orders of magnitudes faster than the prior art. In addition, we have created a large dual focal stack database with ground truth disparity.

Our current implementation limits the input size of our network to be focal stacks of 16 layers. In the future, we will investigate approaches to handle denser focal stack. Further, aside from computer vision, we hope our work will stimulate significant future work in human perception and the nature of human eyes.

%
%
%
%

\bibliographystyle{splncs04}
\bibliography{egbib}

\end{document}